\pdfoutput=1
\documentclass[10pt,twocolumn,letterpaper]{article}

\usepackage{cvpr}
\usepackage{times}
\usepackage{epsfig}
\usepackage{graphicx}
\usepackage{amsmath}
\usepackage{amssymb}
\usepackage{booktabs}
\usepackage{array}
\usepackage{tablefootnote}

\usepackage[pagebackref=true,breaklinks=true,letterpaper=true,colorlinks,bookmarks=false]{hyperref}

\cvprfinalcopy % *** Uncomment this line for the final submission

\def\cvprPaperID{2819} % *** Enter the CVPR Paper ID here

\ifcvprfinal\fi
\begin{document}

%%%%%%%%% TITLE
\title{GeoNet: Unsupervised Learning of Dense Depth, Optical Flow and Camera Pose}

\author{Zhichao Yin~and~Jianping Shi\\
SenseTime Research\\
{\tt\small \{yinzhichao, shijianping\}@sensetime.com}
%\and
%Jianping Shi\\
}

\maketitle
%\thispagestyle{empty}

%%%%%%%%% ABSTRACT
\begin{abstract}
We propose GeoNet, a jointly unsupervised learning framework for monocular depth, optical flow and ego-motion estimation from videos. 
The three components are coupled by the nature of 3D scene geometry, jointly learned by our framework in an end-to-end manner. %, benefiting each other and can be tested independently for corresponding task.
%Building upon recent work dealing with these topics seperately, we design three end-to-end modules in the form of convolutional neural networks with each of them tackling one specific task. 
Specifically, geometric relationships are extracted over the predictions of individual modules and then combined as an image reconstruction loss,  reasoning about static and dynamic scene parts separately. Furthermore, we propose an adaptive geometric consistency loss to increase robustness towards outliers and non-Lambertian regions, which resolves occlusions and texture ambiguities effectively. 
%All of these three modules are trained altogether with fully unsupervised setting, benefiting each other and can be tested independently for corresponding task. 
Experimentation on the KITTI driving dataset reveals that our scheme achieves state-of-the-art results in all of the three tasks, performing better than previously unsupervised methods and comparably with supervised ones.
\end{abstract}

%%%%%%%%% BODY TEXT
\section{Introduction}
\label{sec:intro}
%\jpshi{Logic in this paragraph: Introduce the problem is important. Traditional methods and their drawbacks.}
Understanding 3D scene geometry from video is a fundamental topic in visual perception. It includes many classical computer vision tasks, such as depth recovery, flow estimation, visual odometry, \etc. These technologies have wide industrial applications, including autonomous driving platforms~\cite{chen2015deepdriving}, interactive collaborative robotics~\cite{fong2003survey}, and localization and navigation systems~\cite{fraundorfer2007topological}, \etc.
%\jpshi{Add some applications that geometry understanding is useful, such as autonomous driving, robotics, and others.} 

Traditional \emph{Structure from Motion} (SfM) methods~\cite{newcombe2011dtam, snavely2008modeling} tackle them in an integrated way, which aim to simultaneously reconstruct the scene structure and camera motion. Advances have been achieved recently in robust and discriminative feature descriptors~\cite{bay2008speeded, rublee2011orb}, more efficient tracking systems~\cite{zhang2016efficient}, and better exploitation of semantic level information~\cite{maros16}, \etc. Even though, the proneness to outliers and failure in non-textured regions are still not completely eliminated for their inherent reliance on high-quality low-level feature correspondences.

%\jpshi{Logic in this paragraph: deep learning methods beat traditional one on each aspects, due to its benefits on ?. However it still exists some problems.}
To break through these limitations, deep models~\cite{newell2016stacked, szegedy2015going} have been applied to each of the low-level subproblems and achieve considerable gains against traditional methods. The major advantage comes from big data, which helps capturing high-level semantic correspondences for low level clue learning, thus performing better even in ill-posed regions compared with traditional methods. 
% Mimicing traditional techniques of these tasks, a wide variety of models are adaptively designed in their respective fields. 

Nevertheless, to preserve high performance with more general scenarios, 
large corpus of groundtruth data are usually needed for deep learning. In most circumstances, expensive laser-based setups and differential GPS are required, restricting the data grow to a large scale. Moreover, previous deep models are mostly tailored to solve one specific task, such as depth~\cite{laina2016deeper}, optical flow~\cite{FischerDIHHGSCB15}, camera pose~\cite{kendall2017geometric}, \etc. They do not explore the inherent redundancy among these tasks, which can be formulated by geometry regularities via the nature of 3D scene construction. 

%\jpshi{Logic in this paragraph: Some methods more related to our work, but differs greatly.}
Recent works have emerged to formulate these problems together with deep learning. But %\jpshi{Add some limitations of each one, or the difference between our methods.}, 
all possess certain inherent limitations. For example, they require large quantities of laser scanned depth data for supervision~\cite{UmmenhoferZUMID16}, 
demand stereo cameras as additional equipment for data acquisition~\cite{monodepth17}, 
%involve pre-processing step such as stereo rectification~\cite{monodepth17}, 
or cannot explicitly handle non-rigidity and occlusions~\cite{Vijayanarasimhan17, zhou2017unsupervised}.

\begin{figure} [t]
\begin{center}
   \includegraphics[clip, trim=0cm 0cm 0cm 2.8cm, width=1.0\linewidth]{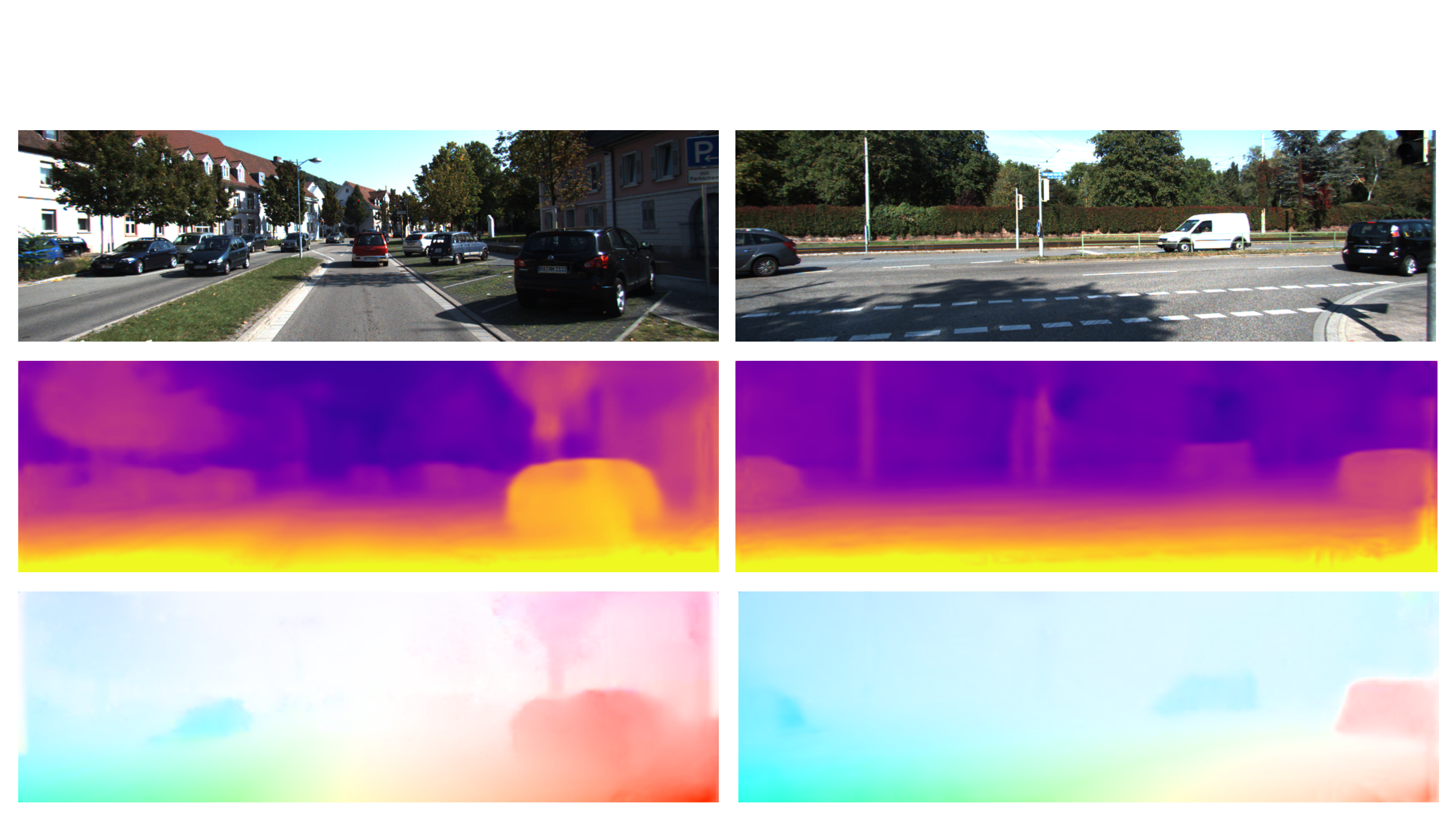}
\end{center}
\vspace{-1ex}
   \caption{Example predictions by our method on KITTI 2015~\cite{Menze2015CVPR}. Top to bottom: input image (one of the sequence), depth map and optical flow. Our model is fully unsupervised and can handle dynamic objects and occlusions explicitly.}
   \vspace{-2ex}
\label{fig::exam}
\end{figure}

%\jpshi{Our methods and advantages of our methods. These paragraph contains too much details and does not highlight your contribution and difference between previous ones. Please reorganize this paragraph}
In this paper, we propose an unsupervised learning framework GeoNet for jointly estimating monocular depth, optical flow and camera motion from video. 
The foundation of our approach is built upon the \emph{nature of 3D scene geometry} (see Sec.~\ref{sec::3dscene} for details). An intuitive explanation is that most of the natural scenes are comprised of rigid staic surfaces, \ie roads, houses, trees,~\etc. Their projected 2D image motion between video frames can be fully determined by the depth structure and camera motion. 
Meanwhile, dynamic objects such as pedestrians and cars commonly exist in such scenes and usually possess the characteristics of large displacement and disarrangement.

As a result, we grasp the above intuition using a deep convolutional network. Specifically, our paradigm employs a divide-and-conquer strategy. A novel cascaded architecture consisting of two stages is designed to solve the scene \emph{rigid flow} and \emph{object motion} adaptively. Therefore the global motion field is able to get refined progressively, making our full learning pipeline a decomposed and easier-to-learn manner. The view synthesis loss guided by such fused motion field leads to natural regularization for unsupervised learning. Example predictions are shown in Fig.~\ref{fig::exam}.

As a second contribution, we introduce a novel adaptive geometric consistency loss to overcome factors not included in a pure view synthesis objective, such as occlusion handling and photo inconsistency issues.
By mimicking the traditional forward-backward (or left-right) consistency check, our approach filters possible outliers and occlusions out automatically. Prediction coherence is enforced between different views in non-occluded regions, while erroneous predictions get smoothed out especially in occluded regions. 

Finally, we perform comprehensive evaluation of our model in all of the three tasks on the KITTI dataset~\cite{Menze2015CVPR}. Our unsupervised approach outperforms previously unsupervised manners and achieves comparable results with supervised ones, which manifests the effectiveness and advantages of our paradigm. Our code will be made available at \url{https://github.com/yzcjtr/GeoNet}.

%To summarize, we propose our contributions upon previous manners as follows:
\iffalse
\begin{itemize}
\item We propose an unsupervised learning framework to jointly tackle depth, optical flow and camera pose estimation. Both rigid and non-rigid cases get adaptively handled by our scheme. The network is trained in an end-to-end fashion without any labeled data, while provides comprehensive understanding of the whole scene geometry by its nature. %\jpshi{xxxx advantages}

\item We develop a novel soft geometric consistency loss which can be applied both in depth and optial flow learning. Coherenct predictions are promised at non-occluded regions while occluded and non-Lambertian cases get properly filtered and smoothed, which in turn significantly improve our final results.% \jpshi{why?}.

\item We conduct comprehensive evaluation of different architecture selection and loss formulation. The result manifests the effectiveness and advantages of our methods.

\end{itemize}
\fi
\vspace{-1ex}

\section{Related Work}
\label{sec:related}
\paragraph{Traditional Scene Geometry Understanding} %3D scene understanding usually consists of factors including depth, camera motion, optical flow, \etc.

Structure-from-Motion (SfM) is a long standing problem which infers scene structure and camera motion jointly from potentially very large unordered image collections~\cite{CGV-052, Hartley2004}. 
%It generally assumes a purely rigid scene and treats non-rigidity as outliers. 
Modern approaches commonly start with feature extraction and matching, followed by geometric verification~\cite{schoenberger2016sfm}. During the reconstruction process, bundle adjustment~\cite{triggs1999bundle} is iteratively applied for refining the global reconstructed structure. Lately wide varieties of methods have been proposed in both global and incremental genres~\cite{sweeney2015optimizing, wu2013towards}. 
%Among them, VisualSFM~\cite{wu2011visualsfm} is a representative framework that elaborates several crafted steps for solving SfM. 
%Recently, researchers work on improving SfM framework by robust and discriminative feature descriptors~\cite{rublee2011orb, bay2008speeded}, more efficient tracking systems~\cite{zhang2016efficient}, and better exploitation of semantic level information~\cite{maros16} \etc. 
However, these existing methods still heavily rely on accurate feature matching. Without good photo-consistency promise, the performance cannot be guaranteed. Typical failure cases may be caused by low texture, stereo ambiguities, occlusions, \etc., which may commonly appear in natural scenes.

% To breakthrough such limitations, researchers recently turn to deep learning methods for substitution of certain submodules \cite{Abhishek17, KendallC17, RoccoAS17}, with the hope of better exploitation of additional information.

Scene flow estimation is another closely related topic to our work, which solves the dense 3D motion field of a scene from stereoscopic image sequences \cite{vedula1999three}. Top ranked methods on the KITTI benchmark typically involve the joint reasoning of geometry, rigid motion and segmentation \cite{Behl2017ICCV, vogel20153d}. %\jpshi{more citations}.
MRFs~\cite{li1994markov} are widely adopted to model these factors as a discrete labeling problem. However, since there exist large quantities of variables to optimize, these off-the-shelf approaches are usually too slow for practical use. On the other hand, several recent methods have emphasized the rigid regularities in generic scene flow. Taniai~\etal~\cite{Taniai2017} proposed to segment out moving objects from the rigid scene with a binary mask. 
Sevilla-Lara~\etal~\cite{sevilla2016optical} defined different models of image motion according to semantic segmentation.
Wulff~\etal~\cite{wulff2017optical} modified the Plane$+$Parallax framework with semantic rigid prior learned by a CNN. 
% Classical optimization techniques are widely adopted including TRW-S~\cite{kolmogorov2006convergent}, MP-PBP~\cite{pacheco2014preserving}, LBFGS~\cite{nocedal1980updating}, \etc.
Different from the above mentioned approaches, we employ deep neural networks for better exploitation of high level cues, not restricted to a specific scenario. Our end-to-end method only takes on the order of milliseconds for geometry inference on a consumer level GPU. %\jpshi{Add more advantages besides speed....}
Moreover, we robustly estimate high-quality ego-motion %regardless of the dynamic objects, 
which is not included in the classical scene flow conception.

\paragraph{Supervised Deep Models for Geometry Understanding}
With recent development of deep learning, great progress has been made in many tasks of 3D geometry understanding, including depth, optical flow, pose estimation, \etc. 

%\jpshi{Reorganize this paragraph by adding one or two single image depth estimation citation. Then the stereo. Then the optical flow.}
%has greatly motivated deep model's application in low-level vision. 
By utilization of a two scale network, Eigen \etal \cite{EigenPF14} demonstrated the capability of deep models for single view depth estimation. While such monocular formulation typically has heavy reliance on scene priors, a stereo setting is preferred by many recent methods. Mayer \etal \cite{MayerIHFCDB15} introduced a correlation layer to mimic traditional stereo matching techniques. Kendall \etal \cite{kendall2017end} proposed 3D convolutions over cost volumes  by deep features to better aggregate stereo information. Similar spirits have also been adopted in learning optical flow. 
E. Ilg \etal \cite{IMKDB17} trained a stacked network on large corpus of synthetic data and achieved impressive result on par with traditional methods.

Apart from the above problems as dense pixel prediction, camera localization and tracking have also proven to be tractable as a supervised learning task. Kendall~\etal~\cite{kendall2015convolutional} cast the 6-DoF camera pose relocalization problem as a learning task, and extended it upon the foundations of multi-view geometry~\cite{kendall2017geometric}. Oliveira~\etal~\cite{OliveiraRBB17} demonstrated how to assemble visual odometry and topological localization modules and outperformed traditional learning-free methods. Brahmbhatt~\etal~\cite{MapNet17} exploited geometric constraints from a diversity of sensory inputs for improving localization accuracy on a broad scale.
%Unfortunately, the acquisition of groundtruth data for above geometry understanding problem is difficult. %Expensive equipments, \ie LiDAR and navigation systems, are usually required. 
%For alleviating such reliances, another direction tends to incorporate geometric contraints in the learning scheme~\cite{MapNet17, kendall2017geometric} besides the supervised loss, which casts light towards fully unsupervised manners.
 
\paragraph{Unsupervised Learning of Geometry Understanding} 

For alleviating the reliances on expensive groundtruth data, various unsupervised approaches have been proposed recently to address the 3D understanding tasks. The core supervision typically comes from a view synthesis objective based on geometric inferences. 
%Recently various unsupervised approaches have been proposed to address the 3D understanding task in more geometrically and economically feasible ways. 
Here we briefly review on the most closely related ones and indicate the crucial differences between ours. 
%\jpshi{You need to re-organize the logic below. No need to explain each method as their abstract. Only highlight its keypoint in unsupervised learning, and difference between your method.}

Garg~\etal~\cite{GargBR16} proposed a stereopsis based auto-encoder for single view depth estimation. While their differentiable inverse warping is based on Taylor expansion, making the training objective sub-optimal.
%Xie \etal \cite{XieGF16} estimated a probabilistic disparity map from a single left image for synthesizing the right view of a binocular pair.
%Instead of inefficiently sampling over all possible disparity range, our method directly predicts depth map (not stereo disparities) and optical flow.
%Guided by the dense motion field, any frame can be synthesized from other frames within the same short sequence without the epipolar constraint.
Both Ren~\etal~\cite{Ren2017UnsupervisedDL} and Yu~\etal~\cite{YuHD16} extended the image reconstruction loss together with a spatial smoothness loss for unsupervised optical flow learning, but took no advantage of geometric consistency among predictions. 
By contrast, Godard~\etal~\cite{monodepth17} exploited such constraints in monocular depth estimation by introducing a left-right consistency loss. 
However, they treat all the pixels equally, which would affect the effectiveness of geometric consistency loss in occluded regions. %(see Sec.~\ref{sec::exp_flow}). 
Concurrent to our work, Meister~\etal~\cite{Meister:2018:UUL} also independently introduce a bidirectional census loss. Different to their stacked structure focusing on unsupervised learning of optical flow, we tackle several geometry understanding tasks jointly. 
Zhou~\etal~\cite{zhou2017unsupervised} mimicked the traditional structure from motion by learning the monocular depth and ego-motion in a coupled way. Building upon the rigid projective geometry, they do not consider the dynamic objects explicitly and in turn learn a explainability mask for compensation. Similarly, Vijayanarasimhan~\etal~\cite{Vijayanarasimhan17} learned several object masks and corresponding rigid motion parameters for modelling moving objects. In contrast, we introduce a residual flow learning module to handle non-rigid cases and emphasize the importance of enforcing geometric consistency in predictions. 

\vspace{-1ex}

\section{Method}
\label{sec:method}

In this section, we start by the nature of 3D scene geometry. Then we give an overview of our GeoNet. It follows by its two components: rigid structure reconstructor and non-rigid motion localizer respectively. Finally, we raise the geometric consistency enforcement, which is the core of our GeoNet.

\begin{figure*}[t]
\begin{center}
   \includegraphics[clip, trim=0cm 0.6cm 0cm 2.8cm, width=1.0\textwidth]{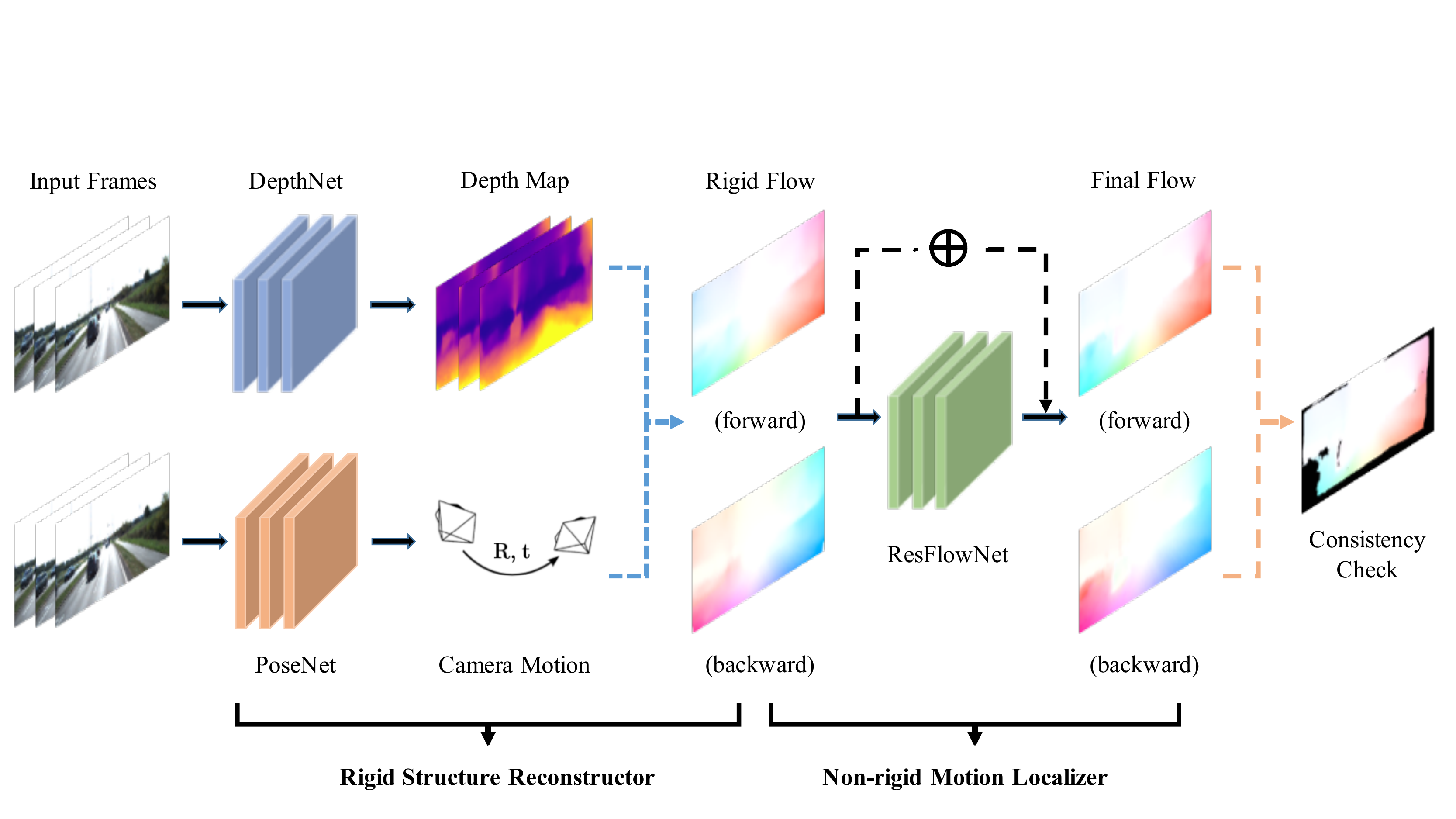}
\end{center}
\vspace{-1ex}
   \caption{Overview of GeoNet. It consists of rigid structure reconstructor for estimating static scene geometry and non-rigid motion localizer for capturing dynamic objects.
   %The inputs to DepthNet and PoseNet differ in that DepthNet takes the sequence as a mini-batch while PoseNet treats them as a single training example. 
   %The rigid flow is computed for any pair of target and source frames bidirectionally, and also treated as independent examples for ResFlowNet. The input to ResFlowNet is not fully depicted here for brevity (see Sec~\ref{sec::resflow} for details). 
   Consistency check within any pair of bidirectional flow predictions is adopted for taking care of occlusions and non-Lambertian surfaces.} % Input sequential frames are concatenated along batch and channel dimension for DepthNet and PoseNet respectively. Our monocular DepthNet predicts the depth maps sequentially, while PoseNet outputs the camera poses of source frames relative to the target frame all at once. Rigid flow is computed from the depth and pose predictions between each pair of target and source frames. Predictions from rigid structure inference (not fully depicted here for brevity, see Sec.~\ref{resflow} for details) are fed into ResFlowNet for learning non-rigid motions. Our final flow builds upon the sum of rigid and residual flow. Moreover, we perform symmetric inference for every pair of frames and check the consistency of predictions (see Sec.~\ref{geocst}). Differentiable warping is performed at both stage and supervises our learning process without any labelled data.} %\jpshi{Add geometric consistency constraint in this figure.}}
   \vspace{-2ex}
\label{fig::pipeline}
\end{figure*}
\subsection{Nature of 3D Scene Geometry}\label{sec::3dscene}

%\jpshi{Describe in this paragraph about the construction of 3D geometry, which includes rigid flow, moving objects in nature. Describe its formulation.}
Videos or images are the screenshots of 3D space projected into certain dimensions. The 3D scene is naturally comprised of static background and moving objects. The movement of static parts in a video is solely caused by camera motion and depth structure. Whereas movement of dynamic objects is more complex, contributed by both homogeneous camera motion and specific object motion.

Understanding the homogeneous camera motion is relatively easier compared to complete scene understanding, since most of the region is bounded by its constraints. %belongs to this motion.
To decompose the problem of 3D scene understanding by its nature, we would like to learn the scene level consistent movement governed by camera motion, namely the \textit{rigid flow}, and the \textit{object motion} separately. %It will in principle make the 3D geometry understanding more a legitimately dissected objective.

Here we briefly introduce the notations and basic concepts used in our paper. To model the strictly restricted rigid flow, we define the static scene geometries by a collection of depth maps $D_i$ for frame $i$, and the relative camera motion $T_{t\to s}$ from target to source frame. The relative 2D rigid flow from target image $I_t$ to source image $I_s$ can be represented by\footnote{Similar to~\cite{zhou2017unsupervised}, we omit the necessary conversion to homogeneous coordinates here for notation brevity.}
\begin{equation}
    \label{equa::proj}
f_{t\to s}^{rig}(p_t) = KT_{t\to s}D_t(p_t)K^{-1}p_t - p_t,
\end{equation}
where $K$ denotes the camera intrinsic and $p_t$ denotes homogeneous coordinates of pixels in frame $I_t$. On the other hand, we model the unconstrained object motion as classical optical flow conception, \ie 2D displacement vectors. We learn the \textit{residual flow} $f^{res}_{t\to s}$ instead of the full representation for non-rigid cases, which we will explain later in Sec.~\ref{sec::resflow}. For brevity, we mainly illustrate the cases from target to source frames in the following, which one can easily generalize to the reversed cases. 
Guided by these positional constraints, we can apply \textit{differentiable inverse warping} \cite{JaderbergSZK15} between nearby frames, which later become the foundation of our fully unsupervised learning scheme. 

\subsection{Overview of GeoNet} \label{sec:overview}

%\jpshi{Overview section should give readers the structure of your framework. Only highlight the name of the modules, and its advantages. So readers could easily grasp the logic of your network. Details should be put into next section.}

Our proposed GeoNet perceives the 3D scene geometry by its nature in an unsupervised manner. In particular, we use separate components to learn the rigid flow and object motion by rigid structure reconstructor and non-rigid motion localizer respectively. The image appearance similarity is adopted to guide the unsupervised learning, which can be generalized to infinite number of video sequences without any labeling cost.% \yinzc{mention geo consistency here?}

An overview of our GeoNet has been depicted in Fig.~\ref{fig::pipeline}. It contains two stages, the rigid structure reasoning stage and the non-rigid motion refinement stage. The first stage to infer scene layout is made up of two sub-networks, \ie the DepthNet and the PoseNet. Depth maps and camera poses are regressed respectively and fused to produce the rigid flow. Furthermore, the second stage is fulfilled by the ResFlowNet to handle dynamic objects. The residual non-rigid flow learned by ResFlowNet is combined with rigid flow, deriving our final flow prediction. Since each of our sub-networks targets at a specific sub-task, the complex scene geometry understanding goal is decomposed to some easier ones. View synthesis at different stage works as fundamental supervision for our unsupervised learning paradigm. 

Last but not the least, we conduct geometric consistency check during training, which significantly enhances the coherence of our predictions and achieves impressive performance.

%The geometric consistency constraints by \jpshi{add to describe the figure, and other advantages of your constraints} are adopted to enable unsupervised learning.
%The depth and relative pose learned by DepthNet and PostNet can directly produce rigid flow. The non-rigid flow learned by the ResFlowNet combined with rigid flow in the first stage leads to our final flow. The ResFlowNet mainly handles the object motion, decomposing the scene understanding problem to a easier one.

\subsection{Rigid Structure Reconstructor}
\label{sec::static}

Our first stage aims to reconstruct the rigid scene structure with robustness towards non-rigidity and outliers. The training examples are temporal continuous frames $I_{i} (i=1\sim n)$ with known camera intrinsics. Typically, a target frame $I_t$ is specified as the reference view, and the other frames are source frames $I_s$. Our DepthNet takes single view as input and exploits accumulated scene priors for depth prediction. During training, 
%Instead of learning multiview stereo, our DepthNet predicts single-view depth by exploiting the accumulated scene cues, which reduces the model complexity in the meantime. 
the entire sequence is treated as a mini-batch of independent images and fed into the DepthNet.
In contrast, to better utilize the feature correspondences between different views, our PoseNet takes the entire sequence concated along channel dimension as input to regress all the relative 6DoF camera poses $T_{t\to s}$ at once. Building upon these elementary predictions, we are able to derive the global rigid flow according to Eq.~\eqref{equa::proj}. Immediately we can synthesize the other view between any pair of target and source frames. Let us denote $\tilde{I}_s^{rig}$ as the inverse warped image from $I_s$ to target image plane by $f_{t\to s}^{rig}$. 
Thereby the supervision signal for our current stage naturally comes in form of minimizing the dissimilarities between the synthesized view $\tilde{I}_s^{rig}$ and original frame $I_t$ (or inversely).

However, it should be pointed out that rigid flow only dominates the motion of non-occluded rigid region while becomes invalid in non-rigid region. Although such negative effect is slightly mitigated within the rather short sequence, we adopt a robust image similarity measurement~\cite{monodepth17} for the photometric loss, which maintains the balance between appropriate assessment of perceptual similarity and modest resilience for outliers, and is differentiable in nature as follows
\begin{equation}
    \label{equa::ssim}
\mathcal{L}_{rw}=\alpha\frac{1-SSIM(I_t,\tilde{I}_s^{rig})}{2}+(1-\alpha)\|I_t-\tilde{I}_s^{rig}\|_1,
\end{equation}
%\begin{split}
%\mathcal{L}_{t\to s}^{rw}=\sum_{p_t\in I_t}\mathcal{C}(I_t,\tilde{I}_s^{rig})(p_t)\\
where SSIM denotes the structural similarity index~\cite{wang2004image} and $\alpha$ is taken to be $0.85$ by cross validation. %\jpshi{Please confirm the notation $s$ $t$ here. Explain each component of the equation not appeared before. Each notation in your paper should have a explanation easy to find nearby for understanding.}
Apart from the rigid warping loss $\mathcal{L}_{rw}$, to filter out erroneous predictions and preserve sharp details, we introduce an edge-aware depth smoothness loss $\mathcal{L}_{ds}$ weighted by image gradients
\begin{equation}
    \label{equa::smooth}
\mathcal{L}_{ds}=\sum_{p_t}|\nabla D(p_t)|\cdot (e^{-|\nabla I(p_t)|})^T,
\end{equation}
where $|\cdot|$ denotes elementwise absolute value, $\nabla$ is the vector differential operator, and T denotes the transpose of image gradient weighting.
%\jpshi{This equation is not clear.}

\subsection{Non-rigid Motion Localizer}
\label{sec::resflow}
%\jpshi{Please reorganize this paragraph since I put description from two sources together.}
The first stage provides us with a stereoscopic perception of rigid scene layout, but ignores the common existence of dynamic objects. Therefore, we raise our second component, \ie the ResFlowNet to localize non-rigid motion. 

Intuitively, generic optical flow can directly model the unconstrained motion, which is commonly adopted in off-the-shelf deep models~\cite{FischerDIHHGSCB15,IMKDB17}. But they do not fully exploit the well-constrained property of rigid regions, which we have already done in the first stage actually. Instead, we formulate our ResFlowNet for learning the \textit{residual non-rigid flow}, the shift solely caused by relative object movement to the world plane. Specifically, we cascade the ResFlowNet after the first stage in a way recommended by \cite{IMKDB17}. For any given pair of frames, the ResFlowNet takes advantage of output from our rigid structure reconstructor, and predicts the corresponding residual signal $f^{res}_{t\to s}$. The final full flow prediction $f^{full}_{t\to s}$ is constituted by $f^{rig}_{t\to s}+f^{res}_{t\to s}$. 

\begin{figure}[t]
\begin{center}
   \includegraphics[clip, trim=0cm 0cm 0cm 2.5cm, width=1.0\linewidth]{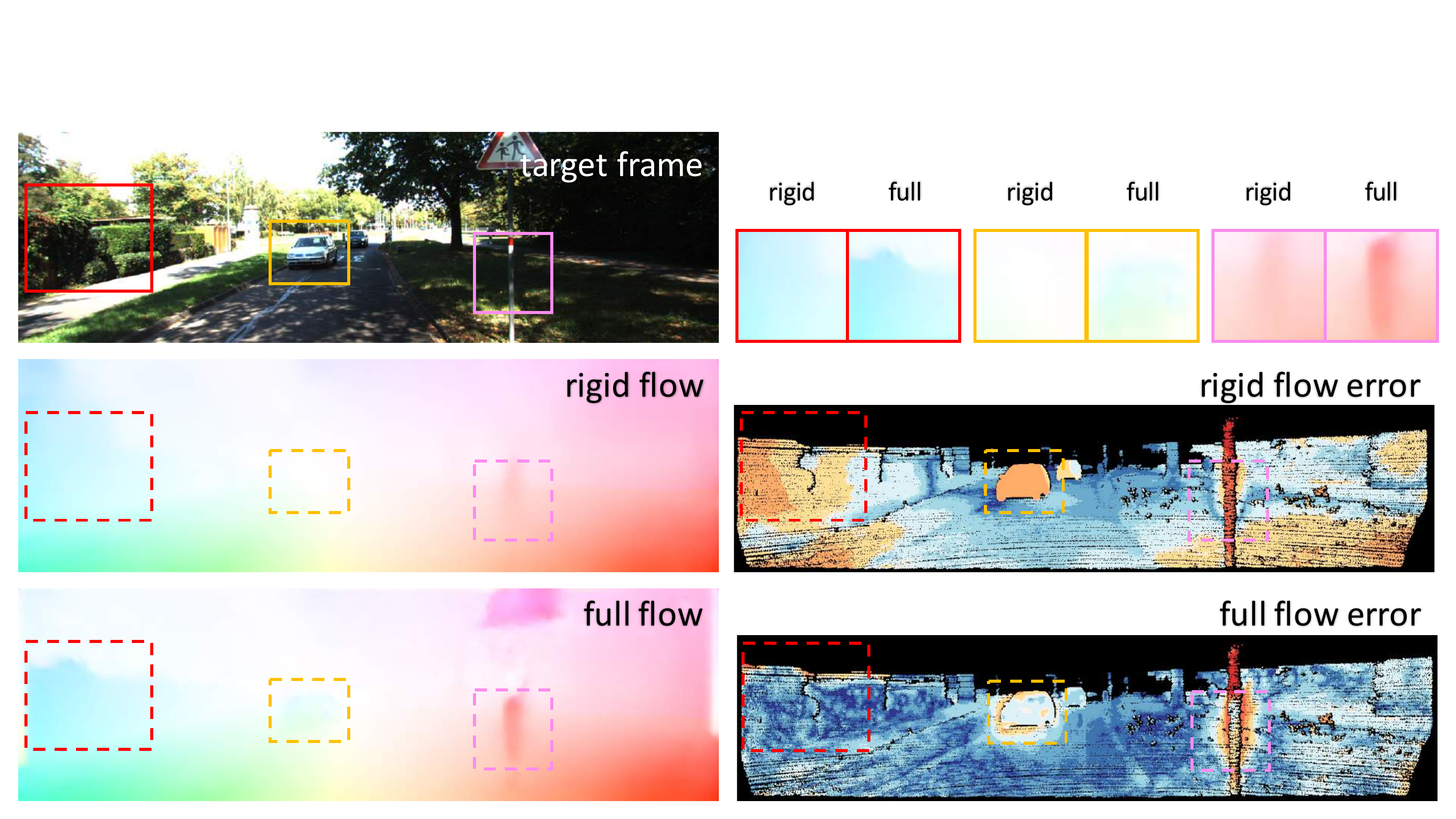}
\end{center}
\vspace{-1ex}
   \caption{Comparison between flow predictions at different stages. Rigid flow gives satisfactory result in most static regions, while residual flow module focuses on localizing non-rigid motion such as cars, and refining initial prediction in challenging cases such as dark illuminations and thin structures.}%\jpshi{Adjust the figure to leave less white space. If possible, highlight some region with closeup.}}
   \vspace{-2ex}
\label{fig::residual}
\end{figure}

%\yinzc{following part placed here or in experiment part?}
As illustrated in Fig.~\ref{fig::residual}, our first stage, rigid structure reconstructor, produces high-quality reconstruction in most rigid scenes, which sets a good starting point for our second stage. Thereby, our ResFlowNet in motion localizer simply focuses on other non-rigid residues. Note that ResFlowNet can not only rectify wrong predictions in dynamic objects, but also refine imperfect results from first stage thanks to our end-to-end learning protocol, which may arise from high saturations and extreme lighting conditions. 
%\yinzc{delete? One may question that since we have an estimation of non-rigidity distribution from second stage, can we apply the related information for guiding the learning of our first stage. Actually we experimented with adjusting the warping loss weight according to the non-rigidty distribution, but found limited improvement. This is perhaps because we have used a robust image similarity measurement, and more importantly, simply down-weighting erroneous image dissimilarity penalty gives no help in rigid regions which occupy the most part of entire scene.} 

Likewise, we can extend the supervision in Sec.~\ref{sec::static} to current stage with slight modifications. In detail, following the full flow $f^{full}_{t\to s}$, we perform image warping between any pair of target and source frames again. Replacing the $\tilde{I}_s^{rig}$ with $\tilde{I}_s^{full}$ in Eq.~\eqref{equa::ssim}, we obtain the full flow warping loss $\mathcal{L}_{fw}$. Similarly, we extend the smoothness loss in Eq.~\eqref{equa::smooth} over 2D optical flow field, which we denote as $\mathcal{L}_{fs}$. %Differently, the dynamic weight %$e^{-|\nabla I(p_t)|}$ in Eq.~\eqref{equa::smooth} is multiplied by $e^{-|\nabla D(p_t)|}$ to ensure uniform depth and optical flow continuity.

\subsection{Geometric Consistency Enforcement}
\label{sec::geocst}
Our GeoNet takes rigid structure reconstructor for static scene, and non-rigid motion localizer as compensation for dynamic objects. Both stages utilize the view synthesis objective as supervision, with the implicit assumption of photometric consistency. Though we employ robust image similarity assessment such as Eq.~\eqref{equa::ssim}, occlusions and non-Lambertian surfaces still cannot be perfectly handled in practice.

%Wrong predictions can easily cheat to achieve optimality during training in above scenarios.
To further mitigate these effects, %traditional methods 
% usually use a robust penalty data term, which we  have already adopted in Eq~\eqref{equa::ssim}. Moreover, 
we apply a forward-backward consistency check in our learning framework without changing the network architecture. The work by Godard \etal \cite{monodepth17} incorporated similar idea into their depth learning scheme %by %simultaneously inferring bidirectional disparities and
with the left-right consistency loss. However, we argue that such consistency constraints, as well as the warping loss, should not be imposed at occluded regions (see Sec.~\ref{sec::exp_flow}). Instead we optimize an adaptive consistency loss across the final motion field.
 
%Specifically, for any pair of frames $I_t$ and $I_s$, our model outputs the bidirectional relative flow $f_{t\to s}$ and $f_{s\to t}$.
%%mapping 2D locations $p_t\in I_t$ to $p_t+f_{t\to s}\in I_s$. 
%The inherent geometric consistency can be imposed by optimizing
%\begin{equation}
%\mathcal{L}_{t\to s}^{fc}=\sum_{p_t\in I_t}|\Delta f_{t\to s}(p_t)|
%\end{equation}
%\iffalse
%\begin{equation}
%\Delta f_{t\to s}(p_t)=f_{s\to t}(p_t+f_{t\to s}(p_t))+f_{t\to s}(p_t)
%\end{equation}
%\fi
%where the $\Delta f_{t\to s}(p_t)$ is the relative error for forward-backward flow consistency check for pixel $p_t$ in $I_t$. \jpshi{What is $^{fc}$ means?}
%%Like the other terms, such loss can be mirrored in a reversed order.
%However, occluded pixels and non-Lambertian surfaces generally violate this condition as there are no corresponding pixels in the other frame. Empirically we find that enforcing such constraint uniformly around the image plane deteriorate the performance in occluded regions. Instead, we propose a soft geometric consistency check with the weight as follows, 
Concretely, our geometric consistency enforcement is fulfilled by optimizing the following objective
\begin{equation}
    \label{equa::cst}
\mathcal{L}_{gc} = \sum_{p_t} [\delta(p_t)]\cdot\|\Delta f_{t\to s}^{full}(p_t)\|_1, 
\end{equation}
where $\Delta f_{t\to s}^{full}(p_t)$ is the full flow difference computed by forward-backward consistency check at pixel $p_t$ in $I_t$, $[\cdot]$ is the Iverson bracket, and $\delta(p_t)$ denotes the condition of  
\begin{equation}
\|\Delta f_{t\to s}^{full}(p_t)\|_2<max\{\alpha, \beta\|f_{t\to s}^{full}(p_t)\|_2\},
%p_{t\to s} (p_t) =1 / (1+e^{\beta \|\Delta f_{t\to s}(p_t)-\alpha\|^2})
\end{equation}
in which $(\alpha,\beta)$ are set to be $(3.0,0.05)$ in our experiment. 
Pixels where the forward/backward flows contradict seriously are considered as possible outliers. Since these regions violate the photo consistency as well as geometric consistency assumptions, we handle them only with the smoothness loss $\mathcal{L}_{fs}$. 
%We would rather employ less but reliable supervision than wrongly import them. 
%The flow predictions in such regions are handled with a smoothness loss as in Eq.~\eqref{equa::smooth}. 
Therefore both our full flow warping loss $\mathcal{L}_{fw}$ and geometric consistency loss $\mathcal{L}_{gc}$ are weighted by $[\delta(p_t)]$ pixelwise.

\iffalse
\begin{equation}
\mathcal{L}_{t\to s}^{fw}=\sum_{p_t\in I_t}P_{t\to s}^{noc}(p_t)\mathcal{C}(I_t,\tilde{I}_s^{fin})(p_t)
\end{equation}

\begin{equation}
\mathcal{L}_{t\to s}^{fc}=\sum_{p_t\in I_t}P_{t\to s}^{noc}(p_t)|\Delta f_{t\to s}(p_t)|
\end{equation}
\fi

\begin{figure*}[t]
\begin{center}
   \includegraphics[clip, trim=0cm 0.2cm 0cm 4.1cm, width=1.0\textwidth]{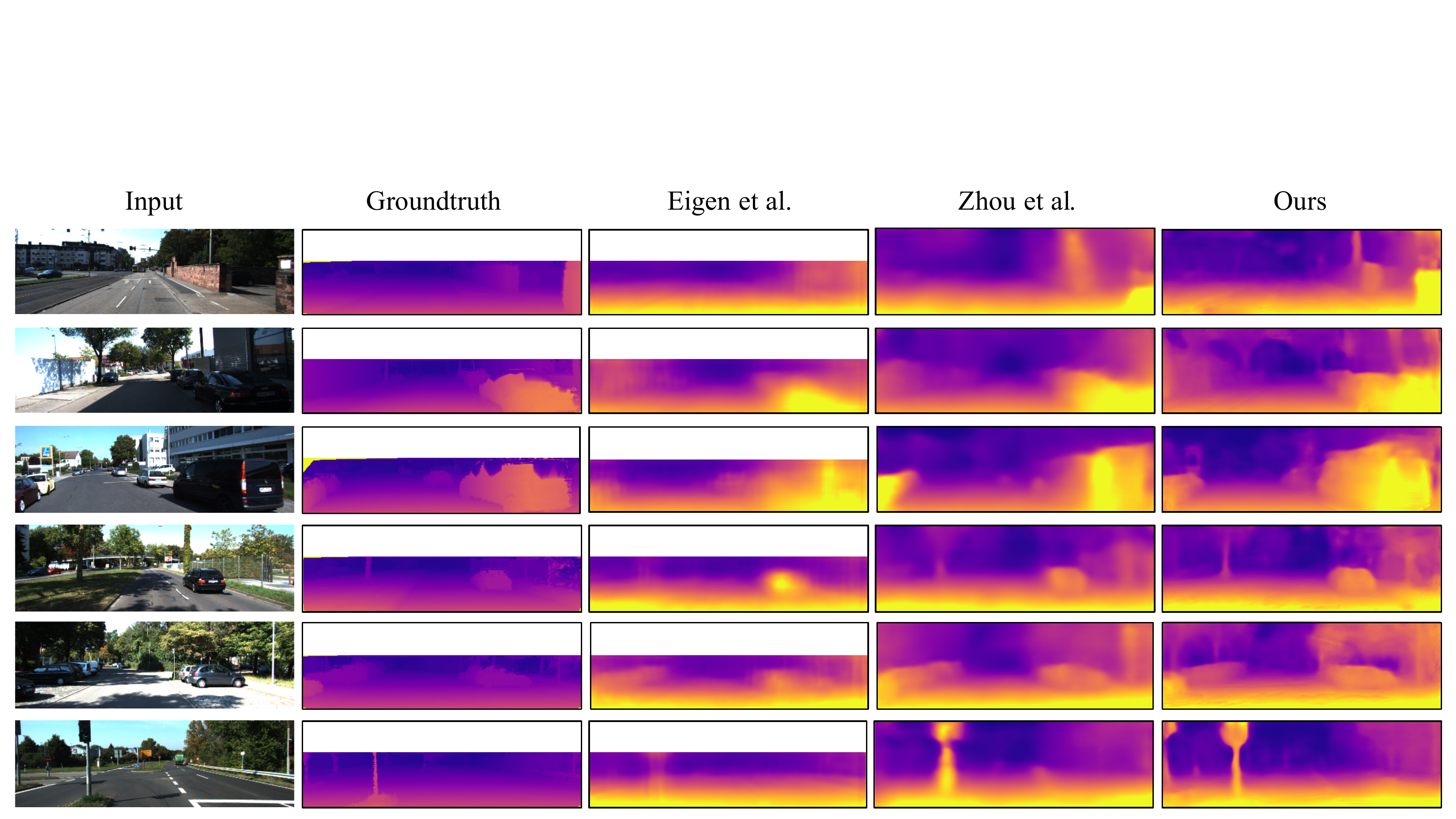}
\end{center}
\vspace{-1ex}
   \caption{Comparison of monocular depth estimation between Eigen~\etal~\cite{EigenPF14} (supervised by depth), Zhou~\etal~\cite{zhou2017unsupervised} (unsupervised) and ours (unsupervised). The groundtruth is interpolated for visualization purpose. Our method captures details in thin structures and preserves consistently high-quality predictions both in close and distant regions.}
   \vspace{-2ex}
\label{fig::exp_depth}
\end{figure*}

%\yinzc{deleted?Note that we mainly discuss about final flow inference above, while simply replacing the final flow with rigid flow, similar strategy can be extended to our first stage to enforce depth prediction's coherence. But we did not see significant improvement and finally excluded it.}

To summarize, our final loss through the entire pipeline becomes
\begin{equation}
\mathcal{L}=\sum_{l}\sum_{\langle t,s \rangle}\{\mathcal{L}_{rw}+\lambda_{ds}\mathcal{L}_{ds}+\mathcal{L}_{fw}+\lambda_{fs}\mathcal{L}_{fs}+\lambda_{gc}\mathcal{L}_{gc}\},
\end{equation}
where $\lambda$ denotes respective loss weight, $l$ indexes over pyramid image scales, and $\langle t,s \rangle$ indexes over all the target and source frame pairs and their inverse combinations. %\jpshi{Please go over all the equations for its annotation, abbreviation}

\vspace{-1ex}

\section{Experiments}
\label{sec:exper}
In this section, we firstly introduce our network architecture and training details. Then we will show qualitative and quantitative results in monocular depth, optical flow and camera pose estimation tasks respectively. %, and Cityscapes for evaluating cross dataset generalization ability.

\subsection{Implementation Details} \label{sec::exp_net}
\paragraph{Network Architecture}
Our GeoNet mainly contains three subnetworks, the DepthNet, the PoseNet, together to form the rigid structure reconstructor, and the ResFlowNet, incorporated with the output from previous stage to localize non-rigid motion. 
Since both the DepthNet and the ResFlowNet reason about pixel-level geometry, we adopt the network architecture in~\cite{monodepth17} as backbone.  
Their structure mainly consists of two components: the encoder and the decoder parts. The encoder follows the basic structure of ResNet50
 as its more effective residual learning manner. The decoder is made up of deconvolution layers to enlarge the spatial feature maps to full scale as input. To preserve both global high-level and local detailed information, we use skip connections between encoder and decoder parts at different corresponding resolutions. %borrowing the merits of hourglass~\cite{newell2016stacked}. 
Both the depth and residual flow are predicted in a multi-scale scheme. %with loss computation in each scale. 
The input to ResFlowNet consists of batches of tensors concated in channel dimension, including the image pair $I_s$ and $I_t$, the rigid flow $f_{t\to s}^{rig}$, the synthesized view $\tilde{I}_s^{rig}$ and its error map compared with original frame $I_t$.
Our PoseNet regresses the  6-DoF camera poses, \ie the euler angles and translational vectors. The architecture is same as in~\cite{zhou2017unsupervised}, which contains 8 convolutional layers followed by a global average pooling layer before final prediction. We adopt batch normalization~\cite{ioffe2015batch} and ReLUs~\cite{nair2010rectified} interlaced with all the convolutional layers except the prediction layers.

\paragraph{Training Details} Our experiment is conducted using the TensorFlow framework~\cite{abadi2016tensorflow}. Though the sub-networks can be trained together in an end-to-end fashion, there is no guarantee that the local gradient optimization could get the network to that optimal point. Therefore, we adopt a stage-wise training strategy, reducing computational cost and memory consumption at meantime. Generally speaking, we first train the DepthNet and the PoseNet, then by fixing their weights, the ResFlowNet is trained thereafter. We also evaluated finetuning the overall network with a smaller batch size and learning rate afterwards, but achieved limited gains. %due to memory bottleneck. 
During training, we resize the image sequences to a resolution of $128\times 416$. We also perform random resizing, cropping, and other color augmentations to prevent overfitting. The network is optimized by Adam~\cite{kingma2014adam}, where $\beta_1=0.9$, $\beta_2=0.999$. 
The loss weights are set to be $\lambda_{ds}=0.5$, $\lambda_{fs}=0.2$ and $\lambda_{gc}=0.2$ for all the experiments.
We take an initial learning rate of 0.0002 and mini-batch size of 4 at both stages. 
The network is trained on a single TitanXP GPU and infers depth, optical flow and camera pose with the speed of $15ms$, $45ms$ and $4ms$ per example at test time. 
The training process typically takes around 30 epochs for the first stage and 200 epochs for the second stage to converge. %, we find the optimization is confronted with slight unstability in the second stage due to the clipping of outliers. 
To make a fair evaluation, we compare our method with different training/test split for each task on the popular KITTI dataset~\cite{Menze2015CVPR}. %\jpshi{The logic here is not clear.}
 
%The PoseNet is a fully convolutional network with 7 conv layers of stride 2. The final prediction is averaged over all spatial outputs. Our PoseNet directly predicts the 6DoF vectors of quaternions and translations, which are converted to transformation matrices for calculating loss.

\begin{table*}[t]
\begin{center}
\small
\setlength{\tabcolsep}{5.0pt}
\begin{tabular*}{1.0\linewidth}{c|c|c|c|c|c|c|c|c|c}
\toprule
Method & Supervised & Dataset & Abs Rel & Sq Rel & RMSE & RMSE log & $\delta<1.25$ & $\delta<1.25^2$ & $\delta<1.25^3$\\
\hline
Eigen~\etal~\cite{EigenPF14} Coarse & Depth & K & 0.214 & 1.605 & 6.563 & 0.292 & 0.673 & 0.884 & 0.957\\
Eigen~\etal~\cite{EigenPF14} Fine & Depth & K & 0.203 & 1.548 & 6.307 & 0.282 & 0.702 & 0.890 & 0.958 \\
Liu~\etal~\cite{liu2016learning} & Depth & K & 0.202 & 1.614 & 6.523 & 0.275 & 0.678 & 0.895 & 0.965\\
Godard~\etal~\cite{monodepth17}  & Pose & K & \bf{0.148} & 1.344 & 5.927 & 0.247 & \bf{0.803} & 0.922 & 0.964\\ 
Zhou~\etal~\cite{zhou2017unsupervised} & No & K & 0.208 & 1.768 & 6.856 & 0.283 & 0.678 & 0.885 & 0.957\\
Zhou~\etal~\cite{zhou2017unsupervised} updated\tablefootnote{\label{git} Results are updated from \url{https://github.com/tinghuiz/SfMLearner} with improved implementation.} & No & K & 0.183 & 1.595 & 6.709 & 0.270 & 0.734 & 0.902 & 0.959\\
Ours VGG & No & K & 0.164 & 1.303 & 6.090 & 0.247 & 0.765 & 0.919 & 0.968 \\
Ours ResNet & No & K & 0.155 & \bf{1.296} & \bf{5.857} & \bf{0.233} & 0.793 & \bf{0.931} & \bf{0.973}\\
\hline
Garg~\etal~\cite{GargBR16} cap 50m & Pose & K & 0.169 & 1.080 & 5.104 & 0.273 & 0.740 & 0.904 & 0.962 \\
Ours VGG cap 50m & No & K & 0.157 & 0.990 & 4.600 & 0.231 & 0.781 & 0.931 & 0.974 \\
Ours ResNet cap 50m & No & K & \bf{0.147} & \bf{0.936} & \bf{4.348} & \bf{0.218} & \bf{0.810} & \bf{0.941} & \bf{0.977}\\
\hline
Godard~\etal~\cite{monodepth17}  & Pose & CS + K & \bf{0.124} & \bf{1.076} & \bf{5.311} & \bf{0.219} & \bf{0.847} & \bf{0.942} & \bf{0.973}\\
Zhou~\etal~\cite{zhou2017unsupervised} & No & CS + K & 0.198 & 1.836 & 6.565 & 0.275 & 0.718 & 0.901 & 0.960 \\
Ours ResNet & No & CS + K & 0.153 & 1.328 & 5.737 & 0.232 & 0.802 & 0.934 & 0.972\\
%Ours ResNet cap 50m & No & CS + K & \bf{0.145} & 0.970 & 4.407 & \bf{0.218} & 0.817 & 0.943 & \bf{0.976} \\
\bottomrule
\end{tabular*}
\end{center}
\caption{Monocular depth results on KITTI 2015~\cite{Menze2015CVPR} by the split of Eigen~\etal~\cite{EigenPF14}. For training, K is the KITTI dataset~\cite{Menze2015CVPR} and CS is Cityscapes~\cite{Cordts2016Cityscapes}. Errors for other methods are taken from~\cite{monodepth17, zhou2017unsupervised}. We show the best result trained only on KITTI in bold. The results of Garg~\etal~\cite{GargBR16} are capped at 50m and we seperately list them for comparison.}
\label{tab:depth}
\vspace{-2ex}
\end{table*}

\subsection{Monocular Depth Estimation} \label{sec::exp_mono}
To evaluate the performance of our GeoNet in monocular depth estimation, we take the split of Eigen~\etal~\cite{EigenPF14} to compare with related works. Visually similar frames to the test scenes as well as static frames are excluded following~\cite{zhou2017unsupervised}.
%Specifically, we remove all the frames covered by the test split with 29 scenes, and take images from the remaining 32 scenes as training data. 
The groundtruth is obtained by projecting the Velodyne laser scanned points into image plane. To evaluate at input image resolution, we resize our predictions by interlinear interpolation. The sequence length is set to be 3 during training. %We perform ablation study with different encoder options. We have also evaluated the experiment setting with first pretraining on Cityscapes dataset~\cite{Cordts2016Cityscapes}, then finetuing on KITTI dataset.

As shown in Table~\ref{tab:depth}, ``Ours VGG'' trained only on KITTI shares the same network architecture with ``Zhou~\etal~\cite{zhou2017unsupervised} without BN'', which reveals the effectiveness of our loss functions. While the difference between ``Ours VGG'' and ``Ours ResNet'' validates the gains achieved by different network architectures.  
%the ResNet based encoder shows clear advantage over the VGG encoder, and pretraining on Cityscapes dataset brings slight improvement. 
Our method significantly outperforms both supervised methods~\cite{EigenPF14, liu2016learning} and previously unsupervised work~\cite{GargBR16, zhou2017unsupervised}. A qualitative comparison has been visualized in Fig.~\ref{fig::exp_depth}. 
Interestingly, our result is slightly inferior to Godard~\etal~\cite{monodepth17} when trained on KITTI and Cityscapes datasets both. We believe this is due to the profound distinctions between training data characteristics,~\ie rectified stereo image pairs and monocular video sequences. 
% which utilizes additional rectified stereo pairs (\ie pose supervision) for training. 
Still, the results manifest the geometry understanding ability of our GeoNet, which successfully captures the regularities among different tasks out of videos.
%The results manifest our GeoNet with depth, pose, and motion jointly inferred together, contributes each other and gets more robust prediction. \jpshi{Please verify the reason here.}

\begin{figure*}
\begin{center}
   \includegraphics[clip, trim=0cm 0cm 0cm 0cm, width=1.0\linewidth]{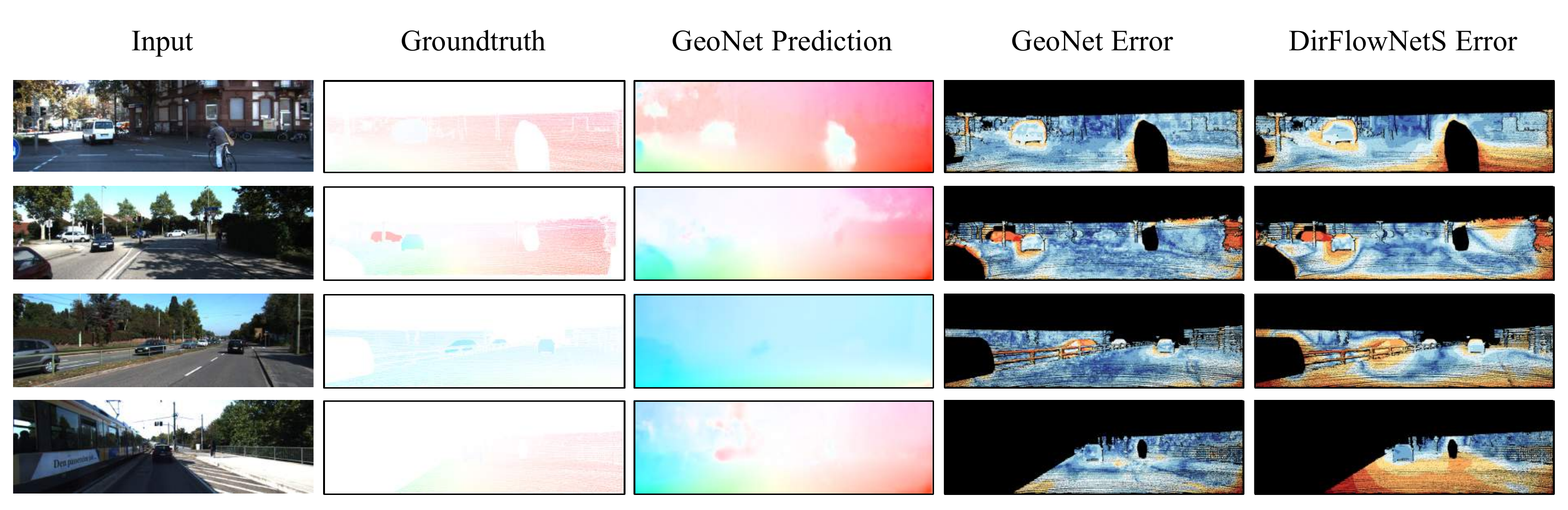}
\end{center}
\vspace{-1ex}
   \caption{Comparison of direct flow learning method DirFlowNetS (geometric consistency loss enforced) and our GeoNet framework. As shown in the figure, GeoNet shows clear advantages in occluded, texture ambiguous regions, and even in shaded dim area.}
   \vspace{-2ex}
\label{fig::exp_flow}
\end{figure*}

\subsection{Optical Flow Estimation} \label{sec::exp_flow}
The performance of optical flow component is validated on the KITTI stereo/flow split. The official 200 training images are adopted as testing set. 
Thanks to our unsupervised nature, we could take the raw images without groundtruth for training. All the related images in the 28 scenes covered by testing data are excluded. 
%Since the optical flow task involves two consecutive frames, we set the sequence length as 2.  
To compare our \textit{residual flow learning} scheme with \textit{direct flow learning}, we specifically trained modified versions of FlowNetS~\cite{FischerDIHHGSCB15} with the unsupervised losses: ``Our DirFlowNetS (no GC)'' is guided by the warping loss and smoothness loss as in Sec.~\ref{sec::resflow}, while ``Our DirFlowNetS'' further incorporates the geometric consistency loss as in Sec.~\ref{sec::geocst} during training.
%warping loss and smoothness loss as in Sec.~\ref{sec::resflow}, denoted as 'Our DirFlowNetS (no GC)'. 
Moreover, we conduct ablation study in adaptive consistency loss versus naive consistency loss,~\ie~without weighting in Eq.~\eqref{equa::cst}.

\begin{table}
\begin{center}
\setlength{\tabcolsep}{5.0pt}
\begin{tabular*}{1.0\linewidth}{c|c|p{1.1cm}<{\centering}|p{1.1cm}<{\centering}}
\toprule
Method & Dataset & Noc & All \\
\hline
EpicFlow~\cite{revaud2015epicflow} & - & 4.45 & 9.57 \\
FlowNetS~\cite{FischerDIHHGSCB15} & C+S & 8.12 & 14.19 \\
FlowNet2~\cite{IMKDB17} & C+T & 4.93 & 10.06 \\
\hline
DSTFlow~\cite{Ren2017UnsupervisedDL} & K & 6.96 & 16.79 \\
%Our Rigid Flow & K & 12.88 &  15.61 \\
Our DirFlowNetS (no GC) & K & 6.80 & 12.86  \\
Our DirFlowNetS & K & \bf{6.77} & 12.21  \\
Our Naive GeoNet & K & 8.57 & 17.18  \\
Our GeoNet & K & 8.05 & \bf{10.81}  \\
\bottomrule
\end{tabular*}
\end{center}
\caption{Average end-point error (EPE) on KITTI 2015 flow training set over non-occluded regions (Noc) and overall regions (All). The handcrafted EpicFlow takes 16s per frame at runtime; The supervised FlowNetS is trained on FlyingChairs and Sintel; Likewise the FlowNet2 is trained on FlyingChairs and FlyingThings3D.}
\label{tab::flow}
\vspace{-3ex}
\end{table}

%\paragraph{Locality of Warping Loss} 
As demonstrated in Table~\ref{tab::flow}, our GeoNet achieves the lowest EPE in overall regions and comparable result in non-occluded regions against other unsupervised baselines. 
The comparison between ``Our DirFlowNetS (no GC)'' and ``Our DirFlowNetS'' already manifests the effectiveness of our geometric consistency loss even in a variant architecture. Futhermore, ``Our GeoNet'' adopts the same losses but beats ``Our DirFlowNetS'' in overall regions, demonstrating the advantages of our architecture based on nature of 3D scene geometry (see Fig.~\ref{fig::exp_flow} for visualized comparison). 
%It demonstrates the effectiveness of our soft geometric consistency enforcement module, especially in occluded area. 
Nevertheless, naively enforcing consistency loss proves to deteriorate accuracy %in occluded regions, 
as shown in ``Our Naive GeoNet'' entry. 

\begin{figure}
\begin{center}
   \includegraphics[width=0.8\linewidth]{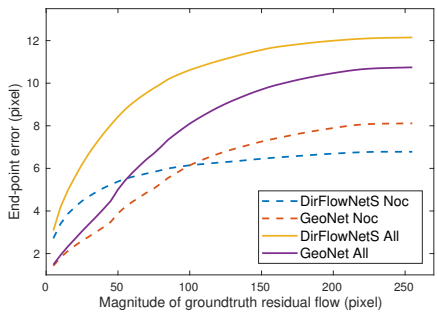}
\end{center}
\vspace{-1ex}
   \caption{Average EPE at different magnitude of groundtruth residual flow. In total regions (All), GeoNet consistently outperforms direct flow regression; but in non-occluded regions (Noc), the advantage of GeoNet is restricted to the neighbourhood of rigid flow.}
   \vspace{-2ex}
\label{fig::exp_flow_error}
\end{figure}

\vspace{-10pt}
\paragraph{Gradient Locality of Warping Loss} However, the direct unsupervised flow network DirFlowNetS performs better in non-occluded regions than GeoNet, which seems unreasonable. We investigate into the end-point error (EPE) distribution over different magnitudes of groundtruth residual flow~\ie $\|f^{gt}-f^{rig}\|$, where $f^{gt}$ denotes the groundtruth full flow. As shown in Fig.~\ref{fig::exp_flow_error}, our GeoNet achieves much lower error in small displacement relative to $f^{rig}$, while the error increases with large displacement. Experimentally, we find that GeoNet is extremely good at rectifying small errors from rigid flow. However, the predicted residual flow tends to prematurely converge to a certain range, which is in consistency with the observations of \cite{monodepth17}. It is because the gradients of warping based loss are derived by local pixel intensity differences, which would be amplified in a more complicated cascaded architecture, ~\ie the GeoNet. We have experimented by replacing the warping loss with a numerically supervised one (guided by groundtruth or knowledge distilled from the DirFlowNetS~\cite{hinton2015distilling}) without changing network architecture, and found such issue disappeared. Investigating practical solution to avoid the gradient locality of warping loss is left as our future work.
%Straightforward solution maybe converting the GeoNet into a supervised variant but contrary to our original intention.

\subsection{Camera Pose Estimation} \label{sec::exp_pose}
We have evaluated the performance of our GeoNet on the official KITTI visual odometry split. %It consists of 22 stereo sequences, among which we take the 11 sequences with groundtruth for validation. 
To compare with Zhou~\etal~\cite{zhou2017unsupervised}, we divide the 11 sequences with groundtruth into two parts: the 00-08 sequences are used for training and the 09-10 sequences for testing. The sequence length is set to be 5 during training. Moreover, we compare our method with a traditional representative SLAM framework: ORB-SLAM~\cite{murAcceptedTRO2015}. It involves global optimization steps such as loop closure detection and bundle adjustment. Here we present two versions: ``The ORB-SLAM (short)'' only takes 5 frames as input and ``ORB-SLAM (long)'' takes the entire sequence as input. All of the results are evaluated in terms of 5-frame trajectories, and scaling factor is optimized to align with groundtruth to resolve scale ambiguity~\cite{sturm2012benchmark}. 
As shown in Table~\ref{tab:pose}, our method outperforms all of the competing baselines. Note that even though our GeoNet only utlizes limited information within a rather short sequence, it still achieves better result than ``ORB-SLAM(full)''. This reveals again that our geometry anchored GeoNet captures additional high level cues other than sole low level feature correspondences. Finally, we analyse the failure cases and find the network sometimes gets confused about the reference system when large dynamic objects appear nearby in front of the camera, which commonly exist in direct visual SLAM~\cite{engel2014lsd}.
%\jpshi{Add a sentence to explain the reason here to echo the advantage of our network design.}

\begin{table}[tbh]
\small
\begin{center}
\setlength{\tabcolsep}{5.0pt}
\begin{tabular*}{1.0\linewidth}{c|c|c}
\toprule
Method & Seq.09 & Seq.10 \\
\hline
ORB-SLAM~(full) & $0.014\pm 0.008$ & $0.012\pm 0.011$\\
ORB-SLAM~(short) & $0.064\pm 0.141$ & $0.064\pm 0.130$\\
Zhou~\etal~\cite{zhou2017unsupervised} & $0.021\pm 0.017$ & $0.020\pm 0.015$\\
Zhou~\etal~\cite{zhou2017unsupervised} updated & $0.016\pm 0.009$ & $0.013\pm 0.009$\\
Our GeoNet &{\bf{ 0.012 $\pm$ 0.007 }}& \bf{0.012 $\pm$ 0.009}\\
\bottomrule
\end{tabular*}
\end{center}
\caption{Absolute Trajectory Error (ATE) on KITTI odometry dataset. The results of other baselines are taken from ~\cite{zhou2017unsupervised}. Our method outperforms all of the other methods.} %even including ORB-SLAM (full).}
\label{tab:pose}
\vspace{-2ex}
\end{table}
\vspace{-1ex}

\section{Conclusion}
\label{sec:conclu}
We propose the jointly unsupervised learning framework GeoNet, and demonstrate the advantages of exploiting geometric relationships over different previously ``isolated'' tasks. Our unsupervised nature profoundly reveals the capability of neural networks in capturing both high level cues and feature correspondences for geometry reasoning. The impressive results compared to other baselines including the supervised ones indicate possibility of learning these low level vision tasks without costly collected groundtruth data.

For future work, we would like to tackle the gradient locality issue of warping based loss, and validate the possible improvement of introducing semantic information into our GeoNet.

\paragraph{Acknowledgements} We would like to thank Guorun Yang and Tinghui Zhou for helpful discussions and sharing the code. We also thank the anonymous reviewers for their instructive comments.
\vspace{-1ex}

{\small
\bibliographystyle{ieee}
\bibliography{egbib}
}

\end{document}